\documentclass[letterpaper]{IEEEtran}
\usepackage{latexsym,,amssymb,amsmath,graphicx,epsf,cite,bbm,float}
\usepackage{ifpdf}
\usepackage{epstopdf}

\usepackage{algorithm,algorithmic}
\usepackage{amsmath,amssymb,bm}
\usepackage{amsfonts,dsfont,color,bbm,subcaption}

\usepackage{amsmath,amsthm}
\usepackage{hyperref}

\newtheorem{theorem}{Theorem}

\newtheorem{lemma}[]{Lemma}

\newtheorem{corollary}[]{Corollary}

\newtheorem{remark}[]{Remark}

\newtheorem{definition}[]{Definition}

\newcommand{\be}{\begin{equation}}
\newcommand{\ee}{\end{equation}}
\newcommand{\bea}{\begin{eqnarray}}
\newcommand{\eea}{\end{eqnarray}}

\newcommand{\MB}{\left[\begin{array}}
\newcommand{\ME}{\end{array}\right]}

\newcommand{\ei}{\end{itemize}}
\newcommand{\bi}{\begin{itemize}}

\newcommand{\abs}[1]{|#1|}

\usepackage{amsmath,amsthm}

\title{Low Regret Binary Sampling Method for Efficient Global Optimization of Univariate Functions}
\author{\IEEEauthorblockN{Kaan Gokcesu}, \IEEEauthorblockN{Hakan Gokcesu} }
\begin{document}
\maketitle

\begin{abstract}
	In this work, we propose a computationally efficient algorithm for the problem of global optimization in univariate loss functions. For the performance evaluation, we study the cumulative regret of the algorithm instead of the simple regret between our best query and the optimal value of the objective function. Although our approach has similar regret results with the traditional lower-bounding algorithms such as the Piyavskii-Shubert method for the Lipschitz continuous or Lipschitz smooth functions, it has a major computational cost advantage. In Piyavskii-Shubert method, for certain types of functions, the query points may be hard to determine (as they are solutions to additional optimization problems). However, this issue is circumvented in our binary sampling approach, where the sampling set is predetermined irrespective of the function characteristics. For a search space of $[0,1]$, our approach has at most $L\log (3T)$ and $2.25H$ regret for $L$-Lipschitz continuous and $H$-Lipschitz smooth functions respectively. We also analytically extend our results for a broader class of functions that covers more complex regularity conditions.
\end{abstract}

\section{Introduction}
	\subsection{Motivation}
	Global optimization aims to minimize the value of a loss function with minimal number of evaluations, which is paramount in a myriad of applications like the calibration of a learning system's hyper-parameter or the design of complex systems \cite{cesa_book, poor_book}. In these kinds of problem scenarios, the performance of a parameter set requires to be evaluated numerically or with cross-validations, which may have a high computational cost. Furthermore, since the search space needs to be sequentially explored, the number of samples to be evaluated needs to be small because of various operational constraints. The evaluation samples may be hard to determine especially if the function does not possess common desirable properties such as convexity or linearity. 
	
	This sequential optimization problem of unknown and possibly non-convex functions are referred to as global optimization \cite{pinter1991global}. It has also been dubbed by the names of derivative-free optimization \cite{rios2013derivative} (since the optimization is done based on solely the function evaluations and the derivatives are inconsequential) and black-box optimization \cite{jones1998efficient}. Over the past years, the global optimization problem has gathered significant attention with various algorithms being proposed in distinct fields of research. It has been studied especially in the fields of non-convex optimization \cite{jain2017non, hansen1991number, basso1982iterative}, Bayesian optimization \cite{brochu2010tutorial}, convex optimization \cite{boyd2004convex,nesterov2003introductory,bubeck2015convex}, bandit optimization \cite{munos2014bandits}, stochastic optimization \cite{shalev2012online, spall2005introduction}; because of its practical applications in distribution estimation \cite{gokcesu2018density,willems,gokcesu2018anomaly,coding2}, multi-armed bandits \cite{neyshabouri2018asymptotically,cesa-bianchi,gokcesu2018bandit}, control theory \cite{tnnls3}, signal processing \cite{ozkan}, game theory \cite{tnnls1}, prediction \cite{gokcesu2016prediction,singer}, decision theory \cite{tnnls4} and anomaly detection \cite{gokcesu2019outlier,gokcesu2016nested,gokcesu2017online}.
	
	\subsection{Related Works}
	Global optimization studies the problem of optimizing a function $f(\cdot)$ with minimal number of evaluations. Generally, only the evaluations of the sampled points are revealed and the information about $f(\cdot)$ or its derivatives are unavailable. When a point $x$ is queried, only its value $f(x)$ is revealed. 

	There exists various heuristic algorithms in literature that deals with this problem such as model-based methods, genetic algorithms and Bayesian optimization. However, most popular approach is still the regularity-based methods since in many applications the system has some inherent regularity with respect to its input-output pair, i.e., $f(\cdot)$ satisfies some regularity condition. Even though some works such as Bartlett et al. \cite{bartlett2019simple} and Grill et al. \cite{grill2015black} use a smoothness regularity in regards to the hierarchical partitioning; traditionally, Lipschitz continuity or smoothness are more common in literature. 
	
	Lipschitz regularity was first studied in Piyavskii's work \cite{piyavskii1972algorithm}, which proposes a sequential deterministic method to solve the global optimization problem. The algorithm works by the iterative construction of a function $F(\cdot)$ that lower bounds the function $f(\cdot)$ and the evaluation of $f(\cdot)$ at a point where $F(\cdot)$ reaches its minimum. In the same year, Shubert has independently published the same algorithm \cite{shubert1972sequential}. Hence, this algorithm has been dubbed the Piyavskii-Shubert algorithm. A lot of research has been done with this algorithm at its base. Jacobsen and Torabi \cite{jacobsen1978global}, Basso \cite{basso1982iterative}, Schoen \cite{schoen1982sequential}, Mayne and Polak \cite{mayne1984outer}, Mladineo \cite{mladineo1986algorithm}, Shen and Zhu \cite{shen1987interval}, Horst and Tuy \cite{horst1987convergence}, Hansen et al. \cite{hansen1991number} study special aspects of its application with examples involving functions that satisfy some Lipschitz condition and propose alternative formulations of the algorithm.
	
	Breiman and Cutler \cite{breiman1993deterministic} proposed a multivariate extension that utilizes the Taylor expansion of $f(\cdot)$ at its core to build the lower bounding function $F(\cdot)$. 
	Baritompa and Cutler \cite{baritompa1994accelerations} propose an alternative acceleration of the Breiman and Cutler’s method. 
	Hansen and Jaumard \cite{hansen1995lipschitz} summarize and discuss the algorithms in literature and present them in an organized and concise way with a high-level computer language. 
	Sergeyev \cite{sergeyev1998global} utilizes smooth auxiliary functions for its construction of the lower bounding function $F(\cdot)$. 
	Ellaia et al. \cite{ellaia2012modified} suggest a variant of Piyavskii-Shubert algorithm that maximizes a univariate differentiable function.
	Brent \cite{brent2013algorithms} proposes a variant where the function $f(\cdot)$ is required to be defined on a compact interval with a bounded second derivative. The work in \cite{gokcesu2021regret} studies the problem with a more generalized Lipschitz regularity.
	For a more detailed perspective, the book of Horst and Tuy \cite{horst2013global} has a general discussion about the role of deterministic algorithms in global optimization.
	
	The performance analysis of optimization algorithms are generally done by the convergence to the optimizer. However, this convergence in the input domain becomes challenging when the objective function does not satisfy nice properties like convexity. For this reason, the convergence is studied with regards to the functional evaluation of the optimizer (i.e., the optimal value of $f(\cdot)$) in global optimization problems.
	
	The study of the number of iterations of Piyavskii-Shubert algorithm was initiated by Danilin \cite{danilin1971estimation}. For its simple regret analysis (which is the evaluation difference between the best estimation so far with the optimal value), a crude regret bound of the form $\tilde{r}_T = O(T^{-1})$ can be obtained from \cite{mladineo1986algorithm} when the function $f(\cdot)$ is Lipschitz continuous. The authors further show that the Piyavskii–Shubert algorithm is minimax optimal and superior to uniform grid search. 
	For univariate functions, the work by Hansen et al. \cite{hansen1991number} derives a bound on the sample complexity for a Piyavskii–Shubert variant, which stops automatically upon returning an $\epsilon$-optimizer. They showed that the number of evaluations required by the algorithm is at most proportional with $\int_0^1(f(x^*)-f(x)+\epsilon)^{-1}dx$. For these results, the authors explicitly study the lower bounding functions and improve upon the results of \cite{danilin1971estimation}. 
	The work by Ellaia et al. \cite{ellaia2012modified} improves upon the previous results of \cite{danilin1971estimation,hansen1991number}.
	The work of Malherbe and Vayatis \cite{malherbe2017global} studies an algorithm called LIPO, which is a variant of the Piyavskii–Shubert algorithm. They obtain upper bounds on the regret rates under strong assumptions. The work of Bouttier et al. \cite{bouttier2020regret} studies the regret bounds of Piyavskii-Shubert algorithm under noisy evaluations.
	Instead of the weaker simple regret, the work in \cite{gokcesu2021regret} provides cumulative regret bounds for variants of Piyavskii-Shubert algorithm. Although \cite{gokcesu2021regret} shows that Piyavskii-Shubert algorithm has nice regret bounds for a variety of different regularity conditions (e.g., Lipschitz continuity and smoothness), the optimization of the lower bounding proxy functions themselves are not always easy to solve. When the operational cost is not only driven by the function evaluation but also selecting queries, alternative approaches are need. In this work, we address this issue and by determining the queries efficiently while also having similar regret performance with the Piyavskii-Shubert algorithm.
	
	\subsection{Contributions and Organization}
	Our contributions and organization are as follows:
	\begin{enumerate}
		\item In \autoref{sec:problem}, we formally define the problem setting and provide a definition for the function to be optimized which also includes the Lipschitz continuous and smooth functions.
		
		\item In \autoref{sec:algorithm}, we provide the binary sampling algorithm, which has computational superiority over Piyavskii-Shubert algorithm over determining the queries.
		
		\item In \autoref{sec:regret}, we study the cumulative regret of the binary sampling algorithm and show that it performs as well as the Piyavskii-Shubert algorithm.
		
		\item In \autoref{sec:examples}, we provide the implementation of the binary sampling algorithm for various types of functions and its corresponding cumulative regret as a function of the number of evaluations $T$.
	
	\end{enumerate}
	
	\section{Problem Definition}\label{sec:problem}
	In this section, we provide the formal problem definition. In the univariate global optimization problem, we want to optimize a function $f(\cdot)$ that maps from a compact set to another such that
	\begin{align}
		f(\cdot) : \Theta\rightarrow\Omega,\label{eq:f}
	\end{align}
	where $\Theta$ and $\Omega$ are compact subsets of the real line $\Re$. Thus, for any $x\in\Theta$, we have $f(x)\in\Omega$.
	However, it is not straightforward to optimize any arbitrary function $f(\cdot)$. To this end, we define a regularity measure. Instead of the restrictive Lipschitz continuity or smoothness \cite{gokcesu2021regret}; we define a weaker, more general regularity condition.
	\begin{definition}\label{def:condition}
		Let the function $f(\cdot)$ that we want to optimize satisfy the following condition:
		\begin{align*}
			|f(x)-f(x_E)|\leq Cd(x-x_E),
		\end{align*}
		where $x_E$ is a local extremum (minimum or maximum) of $f(\cdot)$, $C>0$ is a constant and $d(\cdot)$ is a convex function (metric). 
	\end{definition}
	
	We optimize the function $f(\cdot)$ iteratively by selecting a query point $x_t$ at each time $t$ and receive its evaluation $f(x_t)$. Then, we select the next query point based on the past queries and their evaluations. Hence,
	\begin{align}
		x_{t+1}=\Gamma(x_1,x_2,\ldots,x_{t},f(x_1),f(x_2),\ldots,f(x_{t})),
	\end{align}
	where $\Gamma(\cdot)$ is some function. One such algorithm is the famous Piyavskii–Shubert algorithm (and its variants) \cite{gokcesu2021regret}.
	
	We approach this problem from a loss minimization perspective (as in line with the computational learning theory). We consider the objective function $f(\cdot)$ as a loss function to be minimized by producing query points (predictions) $x_t$ from the compact subset $\Theta$ at each point in time $t$. We define the performance of the predictions $x_t$ for a time horizon $T$ by the cumulative loss incurred up to $T$ instead of the best loss so far at time $T$; i.e., instead of the loss of the best prediction up to time $T$:
	\begin{align}
		\tilde{l}_t \triangleq \min_{t\in\{1,\ldots,T\}}f(x_t),
	\end{align}
	we use the cumulative loss up to time $T$:
	\begin{align}
		L_T\triangleq\sum_{t=1}^{T}f(x_t).
	\end{align}
	Let $x_*$ be a global minimizer of $f(\cdot)$, i.e.,
	\begin{align}
	f(x_*) = \min_{x\in\Theta} f(x).\label{eq:min}
	\end{align}
	As in line with learning theory, we use the notion of regret to evaluate the performance of our algorithm \cite{gokcesu2021regret}. Hence, instead of the simple regret at time $T$:
	\begin{align}
		\tilde{r}_T\triangleq\min_{t\in\{1,\ldots,T\}}f(x_t)-f(x_*),
	\end{align}
	we analyze the cumulative regret up to time $T$:
	\begin{align}
		R_T\triangleq\sum_{t=1}^{T}f(x_t)-\sum_{t=1}^{T}f(x_*)\label{eq:regret}.
	\end{align}
In the next section, we propose our computationally efficient low regret algorithm.
	
	\section{The Binary Sampling Method}\label{sec:algorithm}
	In this section, we design our binary sampling algorithm, which can efficiently optimize the objective function $f(\cdot)$ with low regret bounds. Without loss of generality, we assume $x\in[0,1]$	since for any compact set $\Theta$ as used in \eqref{eq:f}, we can reduce it to the problem of optimization in $[0,1]$ after translating and scaling of the input $x$.
	
	Traditionally, global optimization algorithms (such as Piyavskii–Shubert and its variants) work by creating proxy functions $f^L(\cdot)$ that lower bound the objective function $f(\cdot)$. With each new query at time $t$, the proxy function $f^L(\cdot)$ is updated, i.e., a time varying function $f_t^L(\cdot)$ is created. Each such $f_t^L(\cdot)$ is determined by using the past queries $\{x_\tau\}_{\tau=1}^{t-1}$ and their evaluations $\{f(x_\tau)\}_{\tau=1}^{t-1}$; where each query $x_t$ is selected from the extrema of the proxy $f_t^L(\cdot)$. We observe that the creation of the proxy function itself is inconsequential and the important aspect is determining its extrema. In another perspective, it all boils down to determining some candidate points and their potential evaluations (scores) \cite{gokcesu2021regret}. The framework of the algorithm is given next.
	
	\subsection{Algorithmic Framework}
	The algorithm works as the following.
	\begin{enumerate}
		\item At the start, we sample the boundaries $x=0$, $x=1$; and receive their evaluations $f(0)$, $f(1)$. 
		\item If possible, using $x_0=0, x_1=1, f_0=f(0), f_1=f(1)$ as inputs; we determine a candidate point $x'$ in the set $(x_0,x_1)$ together with its score $s'$, and add it to the potential query list. \label{item:candidate}
		\item From the potential queries, we sample the one with the lowest score; then remove it from the list. Let that query be $x_m$ and its evaluation $f_m=f(x_m)$. Let $x_m$ be the only query between the previous queries $x_l$ and $x_r$ (i.e., in $(x_l,x_r)$) with the corresponding evaluations $f(x_l)=f_l$ and $f(x_r)=f_r$ respectively. \label{item:sample}
		\item We repeat Step \ref{item:candidate} with the inputs: $x_0=x_l$, $x_1=x_m$, $f_0=f(x_l)$, $f_1=f(x_m)$.
		\item We repeat Step \ref{item:candidate} with the inputs: $x_0=x_m$, $x_1=x_r$, $f_0=f(x_m)$, $f_1=f(x_r)$.
		\item We return to Step \ref{item:sample}.
	\end{enumerate}

	\subsection{Determination of the Potential Query \texorpdfstring{$x'$}{x'}}\label{sec:candidate}
	As a design choice, the potential query $x'$ in Step \ref{item:candidate} is selected differently from the lower-bounding algorithms \cite{gokcesu2021regret}.
	As opposed to the point with the lowest possible functional value between the boundaries $x_0$ and $x_1$ (which is the intersection of the lines that pass through points $(x_0,f_0)$, $(x_1,f_1)$ with slopes $-L$, $L$ respectively for Lipschitz continuous functions); we select it as the middle point of $x_0$ and $x_1$. The exact expression is given in the following.
	\begin{definition}\label{thm:candidate}
		For an objective function $f(\cdot)$, given the boundary points $x_0$, $x_1$; the potential query $x'$ between $x_0$ and $x_1$ (i.e., $x'\in(x_0,x_1)$) is given by
		\begin{align*}
			x'=\frac{1}{2}\left(x_1+x_0\right),
		\end{align*}
		which is the middle point in the region $[x_0,x_1]$.
	\end{definition}
	
	\subsection{Determination of the Score \texorpdfstring{$s'$}{s'} of the Candidate Point \texorpdfstring{$x'$}{x'}}\label{sec:score}
	The score $s'$ in Step \ref{item:sample} is a lot more straightforward to determine as opposed to the Piyavskii-Shubert algorithm \cite{gokcesu2021regret}.
	
	\begin{lemma}\label{thm:score}
		For an objective function $f(\cdot)$ satisfying \autoref{def:condition}; given the boundary points $x_0$, $x_1$ and their values $f_0\triangleq f(x_0)$, $f_1\triangleq f(x_1)$, we assign the potential query $x'=\frac{1}{2}\left(x_1+x_0\right)$ the following score
		\begin{align}
			s'=\min(f_0,f_1)-Cd\left(\frac{x_1-x_0}{2}\right),
		\end{align}
		where $C$ and $d(\cdot)$ are as in \autoref{def:condition}, which completely lower bounds the functional evaluation of the region $[x_0,x_1]$.
		
		\begin{proof}
			From the condition, we have
			\begin{align}
				f(x)\geq f_0-Cd(x-x_0),\\
				f(x)\geq f_1-Cd(x_1-x),
			\end{align}
			hence
			\begin{align}
				f(x)\geq \max\left(f_0-Cd(x-x_0),f_1-Cd(x_1-x)\right).
			\end{align}
			We have the following two cases:
			\begin{enumerate}
				\item If $x\leq \frac{x_1-x_0}{2}$:
				\begin{align}
					f(x)\geq& f_0-Cd(x-x_0),\\
					\geq& f_0-Cd\left(\frac{x_1-x_0}{2}\right),\\
					\geq& \min(f_0,f_1)-Cd\left(\frac{x_1-x_0}{2}\right).
				\end{align}
				\item If $x\geq \frac{x_1-x_0}{2}$:
				\begin{align}
					f(x)\geq& f_1-Cd(x-x_0),\\
					\geq& f_1-Cd\left(\frac{x_1-x_0}{2}\right),\\
					\geq& \min(f_0,f_1)-Cd\left(\frac{x_1-x_0}{2}\right),
				\end{align}
			\end{enumerate}
			which concludes the proof.
		\end{proof}
	\end{lemma}
	
	\begin{remark}
		With each new sample, the algorithm iteratively creates new potential query points; and the points not sampled remain unchanged.
	\end{remark}
	\begin{remark}
		Between any two adjacent query points, there will exactly be one potential query, which is their middle. Hence, the number of potential queries grows linearly with the number of queries.
	\end{remark}
\begin{remark}
	Various stopping criterion can be considered; such as stopping the algorithm after a fixed amount of trials or when a sufficient closeness to the optimizer value is reached.
\end{remark}

\begin{remark}
	Even though the algorithm has low computational complexity by design; we can further increase efficiency by eliminating the potential queries with scores that are higher than the minimum function evaluation queried so far. 
\end{remark}

\begin{remark}
	Because of the way the potential queries are determined they can be represented as a binary string which increases memory or communication efficiency.
\end{remark}
	
	\section{General Cumulative Regret Analysis}\label{sec:regret}
	In this section, we study the cumulative regret of the algorithm with the candidate points and scores in \autoref{thm:candidate} and \autoref{thm:score} respectively. Similar to \cite{gokcesu2021regret}, we start by bounding the regret of a single queried point, which will be utilized to derive the cumulative regret.
	\begin{lemma}\label{thm:sampleRegret}
		For an objective function $f(\cdot)$ satisfying \autoref{def:condition}, let $x_m$ be the next sampled query point, which is the middle of the boundary points $x_l$ and $x_r$ together with their corresponding function evaluations $f_l\triangleq f(x_l)$ and $f_r\triangleq f(x_r)$. The regret incurred by the sampling of $x_m$ is bounded as
		\begin{align}
			f(x_m)-\min_{x\in[0,1]}f(x)\leq Cd(x_r-x_l),
		\end{align} 
		where $C$ and $d(\cdot)$ are as in \autoref{def:condition}.
		\begin{proof}
			Since the score $s_m$ of the candidate $x_m$ is a lower bound for the function $f(\cdot)$ between $[x_l,x_r]$, we have
			\begin{align}
				s_m\triangleq \min(f_l,f_r)-Cd\left(\frac{x_r-x_l}{2}\right)\leq \min_{x\in[x_l,x_r]}f(x).
			\end{align}
			Because of the way the algorithm works, each candidate point is inside an interval whose union gives the whole domain. Since we sample the candidate point with the lowest score, we have
			\begin{align}
				s_m\triangleq\min(f_l,f_r)-Cd\left(\frac{x_r-x_l}{2}\right)\leq \min_{x\in[0,1]}f(x).
			\end{align}
			Without loss of generality let $f_l\leq f_r$. Then, because of the continuity condition, we have
			\begin{align}
				f(x_m)\leq& f_l+Cd(x'-x_l)-Cd(x'-x_m),
			\end{align}
			for some $x'$ (where $x_r\geq x'\geq x_m$), which satisfies
			\begin{align}
				f_l+Cd(x'-x_l)=f_r+Cd(x_r-x').
			\end{align}
			Thus,
			\begin{align}
				f(x_m)-\min_{x\in[0,1]}f(x)\leq& Cd(x'-x_l)-Cd(x'-x_m)\\
				&+Cd(x_m-x_l).				
			\end{align}
			Since $d(\cdot)$ is convex, we have
			\begin{align}
				f(x_m)-\min_{x\in[0,1]}f(x)\leq& Cd(x_r-x_l)-Cd(x_r-x_m)\\
				&+Cd(x_m-x_l),\\
				\leq& Cd(x_r-x_l),				
			\end{align}
			which concludes the proof.
		\end{proof}
	\end{lemma}
	This result bounds the individual regret of a sampled point $x_m$ with only its boundary values $x_l$ and $x_r$ (irrespective of the functional evaluations $f_l$ and $f_r$), hence, is a worst case bound. 
	
	\begin{remark}
		The result of \autoref{thm:sampleRegret} shows that the regret of our algorithm is linearly dependent with the constant $C$.
	\end{remark}

	Next, we derive the cumulative regret bound up to time horizon $T$.

	\begin{lemma}\label{thm:regret}
		The algorithm has the following regret
		\begin{align*}
			R_T\leq Cd(1/2)+C\sum_{i=0}^{a-1}2^id\left(\frac{1}{2^i}\right)+CBd\left(\frac{1}{2^a}\right), 
		\end{align*}
		where $a$ and $B$ such that $1\leq B \leq 2^a$ and $2^a+B+1=T$.
		\begin{proof}
			We run the algorithm for $T$ sampling times. After sampling the boundary points $x=0$, $x=1$ and the middle point $x=1/2$, there exists $T-3$ more sampling. For a boundary point pair $(x_l,x_r)$, we sample the middle point
			\begin{align}
				x_m\triangleq\frac{x_l+x_r}{2}
			\end{align}
			and incur the individual regret
			\begin{align}
				f(x_m)-\min_{x\in[0,1]}f(x)\leq Cd(x_r-x_l).
			\end{align}
			The distance between any new boundary points will be halved. Hence, at the worst case scenario, we will sample the oldest created candidate points first. The total regret for $T-3$ sampling will be given by
			\begin{align}
				\tilde{R}_T\leq& 2Cd\left(\frac{1}{2}\right)+4Cd\left(\frac{1}{4}\right)+\ldots\\
				&+2^{a-1}Cd\left(\frac{1}{2^{a-1}}\right)+BCd\left(\frac{1}{2^a}\right),
			\end{align}
			where
			\begin{align}
				1\leq B \leq 2^a,
			\end{align}
			and
			\begin{align}
				\sum_{i=1}^{a-1}2^i=&2^a-2,\\
				=&T-3-B.
			\end{align}
			The regret incurred by the first three samples $x=0$, $x=1/2$ and $x=1$ is given by
			\begin{align}
				f(0)+&f\left(\frac{1}{2}\right)+f(1)-3\min_{x\in[0,1]}f(x)\nonumber\\
				&\leq Cd(x_*)+Cd\left(x_*-\frac{1}{2}\right)+Cd(x_*-1),\\
				&\leq Cd\left(\frac{1}{2}\right)+Cd(1),
			\end{align}
			because of the convexity of $d(\cdot)$ from \autoref{def:condition}. Hence, the total regret is
			\begin{align}
				R_T\leq Cd(1/2)+C\sum_{i=0}^{a-1}2^id\left(\frac{1}{2^i}\right)+CBd\left(\frac{1}{2^a}\right),
			\end{align}
			which concludes the proof.
		\end{proof}
	\end{lemma}

	This result shows that the cumulative regret is strongly related with the distance function $d(\cdot)$. In the next section, we provide various examples of this function (including the ones in \cite{gokcesu2021regret}) together with the algorithmic implementation and the regret results.
	
	\section{Results for Various Functions Classes}\label{sec:examples}
	\subsection{Lipschitz Continuous Functions}\label{sec:Lcont}
	We say the function $f(\cdot)$ is Lipschitz continuous with $L$ if 
	\begin{align}
	\abs{f(x)-f(y)}\leq L\abs{x-y},\label{eq:Lipschitz}
	\end{align}
	which implies the absolute of its derivative is bounded from above by $L$, i.e.,
	\begin{align}
	-L\leq f'(x)\leq L.
	\end{align}
	From \eqref{eq:Lipschitz}, we can see that for any extremum $x_E$ of $f(\cdot)$, we have
	\begin{align}
	|f(x)-f(x_E)|\leq L|x-x_E|.
	\end{align}
	This condition is a special case of the more general condition in \autoref{def:condition}, where the constant $C$ is $L$ and the convex distance function $d(\cdot)$ is the absolute function $|\cdot|$, i.e.,
	\begin{align}
		C\rightarrow& L,\label{CL}\\
		d(\cdot)\rightarrow& |\cdot|\label{dL}.
	\end{align}
	
	For this problem setting, the algorithmic implementation (i.e., the score selection) is done as the following.
	\begin{corollary}\label{thm:scoreL}
		For an $L$-Lipschitz continuous function $f(\cdot)$, given the boundary points $x_0$, $x_1$ and their values $f_0\triangleq f(x_0)$ and $f_1\triangleq f(x_1)$, we assign the 'candidate' point $x'=\frac{1}{2}\left(x_1+x_0\right)$ the following score
		\begin{align*}
			s'=\min(f_0,f_1)-L\left|\frac{x_1-x_0}{2}\right|,
		\end{align*}
		which completely lower bounds the functional evaluation of the region $[x_0,x_1]$.
		
		\begin{proof}
			The proof comes from \autoref{thm:score} using \eqref{CL} and \eqref{dL}.
		\end{proof}
	\end{corollary}
	
	Next, we provide the cumulative regret for Lipschitz continuous functions.
	\begin{theorem}\label{thm:regretL}
		The algorithm has the following regret
		\begin{align*}
			R_T \leq L\log(3T).
		\end{align*}
	\begin{proof}
		Using \autoref{thm:regret} together with \eqref{CL} and \eqref{dL}, we get the following:
		\begin{align}
			R_T\leq \frac{L}{2}+L\sum_{i=0}^{a-1}2^i\frac{1}{2^i}+LB\frac{1}{2^a},
		\end{align}
		where $a$ and $B$ such that $1\leq B \leq 2^a$ and $2^a+B+1=T$. Thus,
		\begin{align}
			R_T\leq& \frac{L}{2}+La+L,\\
			\leq& L\log T+L\frac{3}{2},\\
			\leq& L\log(3T),
		\end{align}
		which concludes the proof.
	\end{proof}
	\end{theorem}

	Interestingly, the binary sampling algorithm performs twice as better in comparison to the lower bounding Piyavskii-Shubert algorithm, which has a regret of $\approx2L\log T$ \cite{gokcesu2021regret}.
	
	\subsection{Lipschitz Smooth Functions}\label{sec:Lsmooth}
	In this section, we aim to minimize a Lipschitz smooth function $f(\cdot)$. We say the function $f(\cdot)$ is Lipschitz smooth with $H$ if 
	\begin{align}
		\abs{f'(x)-f'(y)}\leq 2H\abs{x-y},\label{eq:Smooth}
	\end{align}
	which implies the absolute of its second derivative is bounded from above by $H$, i.e.,
	\begin{align}
		-2H\leq f''(x)\leq 2H.
	\end{align} 
	From \eqref{eq:Lipschitz}, we can see that for any extremum $x_E$ of $f(\cdot)$, we have
	\begin{align}
		|f(x)-f(x_E)|\leq H|x-x_E|^2.
	\end{align}
	This condition is a special case of the more general condition in \autoref{def:condition}, where the constant $C$ is $H$ and the convex distance function $d(\cdot)$ is the squared function $|\cdot|^2$, i.e.,
	\begin{align}
		C\rightarrow& H,\label{CH}\\
		d(\cdot)\rightarrow& |\cdot|^2\label{dH}.
	\end{align}
	
	For this problem setting, the algorithmic implementation (i.e., the score selection) is done as the following.
	\begin{corollary}\label{thm:scoreH}
		For an $H$-Lipschitz smooth function $f(\cdot)$, given the boundary points $x_0$, $x_1$ and their values $f_0\triangleq f(x_0)$ and $f_1\triangleq f(x_1)$, the 'candidate' point $x'=\frac{1}{2}\left(x_1+x_0\right)$ is assigned the following score:
		\begin{align*}
			s'=\min(f_0,f_1)-H\left|\frac{x_1-x_0}{2}\right|^2,		
		\end{align*}
		which completely lower bounds the functional evaluation of the region $[x_0,x_1]$.
		\begin{proof}
			The proof comes from \autoref{thm:score} using \eqref{CH} and \eqref{dH}.
		\end{proof}
	\end{corollary}
	
	Next, we provide the cumulative regret for Lipschitz smooth functions.
	\begin{theorem}\label{thm:regretH}
		The algorithm has the following regret
		\begin{align*}
			R_T \leq 2.25H .
		\end{align*}
		\begin{proof}
			Using \autoref{thm:regret} together with \eqref{CH} and \eqref{dH}, we get the following:
			\begin{align}
				R_T\leq \frac{H}{4}+H\sum_{i=0}^{a-1}2^i\frac{1}{2^{2i}}+HB\frac{1}{2^{2a}},
			\end{align}
			where $a$ and $B$ such that $1\leq B \leq 2^a$ and $2^a+B+1=T$. Thus,
			\begin{align}
				R_T\leq& \frac{H}{4}+H\sum_{i=0}^{a}2^i\frac{1}{2^{2i}},\\
				\leq& \frac{H}{4}+H\sum_{i=0}^{\infty}\frac{1}{2^{i}},\\
				\leq& 2.25H.
			\end{align}
			which concludes the proof.
		\end{proof}
	\end{theorem}

	This time the binary sampling algorithm has a regret overhead as expected in comparison to the lower bounding algorithm, which has a regret of $2H$ \cite{gokcesu2021regret}.
	
	\subsection{Polynomial Extension}\label{sec:Lanalytic}
	In this section, we extend the results in \autoref{sec:Lcont} and \autoref{sec:Lsmooth} to a broader class of objective functions by optimizing a function $f(\cdot)$ which satisfy the following condition
	\begin{align}
		|f(x)-f(x_E)|\leq K|x-x_E|^p,\label{eq:analytic}
	\end{align}
	where $x_E$ is a local extremum (minimum or maximum) of $f(\cdot)$, $K>0$ is a constant and $1\leq p$. 
	Note that,
	\begin{itemize}
		\item When $p=1$, this class of functions include the Lipschitz continuous functions in \autoref{sec:Lcont}.
		\item When $p=2$, this class of functions include the Lipschitz smooth functions in \autoref{sec:Lsmooth}.
	\end{itemize} 
	This condition is a special case of the more general one in \autoref{def:condition}, where the constant $C$ is $K$ and the convex function $d(\cdot)$ is the $p^{th}$-order absolute function $|\cdot|^p$, i.e.,
	\begin{align}
		C\rightarrow& K,\label{CK}\\
		d(\cdot)\rightarrow& |\cdot|^p\label{dK}.
	\end{align}
	
	For this problem setting, the algorithmic implementation (i.e., the score selection) is done as the following.
	\begin{corollary}\label{thm:scoreK}
		For a function $f(\cdot)$ satisfying \eqref{eq:analytic}, given the boundary points $x_0$, $x_1$ and their values $f_0\triangleq f(x_0)$ and $f_1\triangleq f(x_1)$, the 'candidate' point $x'=\frac{x_1+x_0}{2}$ has the score
		\begin{align}
			s'=\min(f_0,f_1)-K\left|\frac{x_1-x_0}{2}\right|^p.,		\end{align}
		which lower bounds the functional evaluation of $[x_0,x_1]$.
		\begin{proof}
			The proof comes from \autoref{thm:score} using \eqref{CK}, \eqref{dK}.
		\end{proof}
	\end{corollary}
	
	\begin{theorem}\label{thm:regretK}
		The algorithm has the following regret
		\begin{align*}
			R_T \leq& \frac{K}{2^p}+K\frac{1-2^{(1-p)\log(2T)}}{1-2^{1-p}}.
		\end{align*}
		\begin{proof}
			Using \autoref{thm:regret} together with \eqref{CK} and \eqref{dK}, we get
			\begin{align}
				R_T\leq \frac{K}{2^p}+K\sum_{i=0}^{a-1}2^i\frac{1}{2^{pi}}+KB\frac{1}{2^{pa}},
			\end{align}
			where $a$, $B$ such that $1\leq B \leq 2^a$, $2^a+B+1=T$. Thus,
			\begin{align}
				R_T\leq& \frac{K}{2^p}+K\sum_{i=0}^{a}2^i\frac{1}{2^{pi}},\\
				\leq& \frac{K}{2^p}+K\frac{1-2^{(1-p)(a+1)}}{1-2^{1-p}},\\
				\leq& \frac{K}{2^p}+K\frac{1-2^{(1-p)\log(2T)}}{1-2^{1-p}},
			\end{align}
			which concludes the proof.
		\end{proof}
	\end{theorem}
	
	\begin{itemize}
		\item When $p=1$, $K=L$ as in \autoref{sec:Lcont}, we have the same result as in \autoref{thm:regretL} from L'Hospital.
		
		\item When $p=2$, $K=H$ as in \autoref{sec:Lsmooth}, we have the same result as in \autoref{thm:regretH} from $T< \infty$.
		
		\item When $p\geq2$, choosing $T\rightarrow \infty$ gives
		\begin{align*}
			R_T\leq \left(1+\frac{1}{2^p}+\frac{1}{2^{p-1}-1}\right)K,
		\end{align*}
		for any $p\geq 2$. Hence, $R_T\leq 2.25K$ for all $p\geq 2$.
	\end{itemize}

	As expected, the binary sampling has a $K2^{-p}$ regret overhead in comparison to the lower bounding algorithm \cite{gokcesu2021regret}.

	\subsection{Convex Extension}\label{sec:Lgeneral}
In this section, we extend the results in \autoref{sec:Lcont}, \autoref{sec:Lsmooth} and \autoref{sec:Lanalytic} to a general class of convex functions. 

Here, we aim to optimize a function $f(\cdot)$ which satisfy the following condition
\begin{align}
	|f(x)-f(x_E)|\leq Mg(x-x_E),\label{eq:general}
\end{align}
where $x_E$ is a local extremum (minimum or maximum) of $f(\cdot)$, $M>0$ is a constant and $g(\cdot)$ is non-negative convex and continuous with $g(0)=0$. 

This condition is the case of \autoref{def:condition}, where the constant $C$ is $M$ and the convex distance function $d(\cdot)$ is a general convex function $g(\cdot)$, i.e.,
\begin{align}
	C\rightarrow& M,\label{CG}\\
	d(\cdot)\rightarrow& g(\cdot)\label{dG}.
\end{align}

For this problem setting, the algorithmic implementation (i.e., the score selection) is done as the following.
\begin{corollary}\label{thm:scoreG}
	For a function $f(\cdot)$ satisfying \eqref{eq:general}, given the boundary points $x_0$, $x_1$ and their values $f_0\triangleq f(x_0)$ and $f_1\triangleq f(x_1)$, the 'candidate' point $x'=\frac{x_1+x_0}{2}$ has the score
	\begin{align}
		s'=\min(f_0,f_1)-Mg\left(\frac{x_1-x_0}{2}\right).,		\end{align}
	which completely lower bounds the functional evaluation of the region $[x_0,x_1]$.
	\begin{proof}
		The proof comes from \autoref{thm:score} using \eqref{CG} and \eqref{dG}.
	\end{proof}
\end{corollary}

Next, we provide the cumulative regret.
\begin{theorem}\label{thm:regretG}
	The algorithm has the following regret
	\begin{align*}
		R_T\leq& Mg(1)\log(3T),
	\end{align*}
	\begin{proof}
		Using \autoref{thm:regret} together with \eqref{CG} and \eqref{dG}, we get the following:
		\begin{align}
			R_T\leq Mg\left(\frac{1}{2}\right)+M\sum_{i=0}^{a-1}2^ig\left(\frac{1}{2^{i}}\right)+MBg\left(\frac{1}{2^{a}}\right),
		\end{align}
		where $a$ and $B$ such that $1\leq B \leq 2^a$ and $2^a+B+1=T$. Thus,
		\begin{align}
			R_T\leq& Mg\left(\frac{1}{2}\right)+M\sum_{i=0}^{a}2^ig\left(\frac{1}{2^{i}}\right),\\
			\leq& \frac{1}{2}Mg\left({1}\right)+(a+1)Mg(1),
		\end{align}
		because of convexity. Hence,
		\begin{align}
			R_T\leq& (a+1.5)Mg\left({1}\right),\\
			\leq& Mg(1)\log(3T),
		\end{align}
		which concludes the proof.
	\end{proof}
\end{theorem}

	For more general regularity conditions, it is not straightforward to derive regret bounds for the lower bounding algorithms \cite{gokcesu2021regret}. Nonetheless, the result of binary sampling is intuitive, since the absolute function is the limit of convexity.


\bibliographystyle{ieeetran}
\bibliography{double_bib}

\begin{thebibliography}{10}
\providecommand{\url}[1]{#1}
\csname url@samestyle\endcsname
\providecommand{\newblock}{\relax}
\providecommand{\bibinfo}[2]{#2}
\providecommand{\BIBentrySTDinterwordspacing}{\spaceskip=0pt\relax}
\providecommand{\BIBentryALTinterwordstretchfactor}{4}
\providecommand{\BIBentryALTinterwordspacing}{\spaceskip=\fontdimen2\font plus
\BIBentryALTinterwordstretchfactor\fontdimen3\font minus
  \fontdimen4\font\relax}
\providecommand{\BIBforeignlanguage}[2]{{%
\expandafter\ifx\csname l@#1\endcsname\relax
\typeout{** WARNING: IEEEtran.bst: No hyphenation pattern has been}%
\typeout{** loaded for the language `#1'. Using the pattern for}%
\typeout{** the default language instead.}%
\else
\language=\csname l@#1\endcsname
\fi
#2}}
\providecommand{\BIBdecl}{\relax}
\BIBdecl

\bibitem{cesa_book}
N.~Cesa-Bianchi and G.~Lugosi, \emph{Prediction, learning, and games}.\hskip
  1em plus 0.5em minus 0.4em\relax Cambridge university press, 2006.

\bibitem{poor_book}
H.~V. Poor, \emph{An Introduction to Signal Detection and Estimation}.\hskip
  1em plus 0.5em minus 0.4em\relax NJ: Springer, 1994.

\bibitem{pinter1991global}
J.~D. Pint{\'e}r, ``Global optimization in action,'' \emph{Scientific
  American}, vol. 264, pp. 54--63, 1991.

\bibitem{rios2013derivative}
L.~M. Rios and N.~V. Sahinidis, ``Derivative-free optimization: a review of
  algorithms and comparison of software implementations,'' \emph{Journal of
  Global Optimization}, vol.~56, no.~3, pp. 1247--1293, 2013.

\bibitem{jones1998efficient}
D.~R. Jones, M.~Schonlau, and W.~J. Welch, ``Efficient global optimization of
  expensive black-box functions,'' \emph{Journal of Global optimization},
  vol.~13, no.~4, pp. 455--492, 1998.

\bibitem{jain2017non}
P.~Jain and P.~Kar, ``Non-convex optimization for machine learning,''
  \emph{Foundations and Trends{\textregistered} in Machine Learning}, vol.~10,
  no. 3-4, pp. 142--336, 2017.

\bibitem{hansen1991number}
P.~Hansen, B.~Jaumard, and S.-H. Lu, ``On the number of iterations of
  piyavskii's global optimization algorithm,'' \emph{Mathematics of Operations
  Research}, vol.~16, no.~2, pp. 334--350, 1991.

\bibitem{basso1982iterative}
P.~Basso, ``Iterative methods for the localization of the global maximum,''
  \emph{SIAM Journal on Numerical Analysis}, vol.~19, no.~4, pp. 781--792,
  1982.

\bibitem{brochu2010tutorial}
E.~Brochu, V.~M. Cora, and N.~De~Freitas, ``A tutorial on bayesian optimization
  of expensive cost functions, with application to active user modeling and
  hierarchical reinforcement learning,'' \emph{arXiv preprint arXiv:1012.2599},
  2010.

\bibitem{boyd2004convex}
S.~Boyd, S.~P. Boyd, and L.~Vandenberghe, \emph{Convex optimization}.\hskip 1em
  plus 0.5em minus 0.4em\relax Cambridge university press, 2004.

\bibitem{nesterov2003introductory}
Y.~Nesterov, \emph{Introductory lectures on convex optimization: A basic
  course}.\hskip 1em plus 0.5em minus 0.4em\relax Springer Science \& Business
  Media, 2003, vol.~87.

\bibitem{bubeck2015convex}
S.~Bubeck, ``Convex optimization: Algorithms and complexity,''
  \emph{Foundations and Trends{\textregistered} in Machine Learning}, vol.~8,
  no. 3-4, pp. 231--357, 2015.

\bibitem{munos2014bandits}
R.~Munos, ``From bandits to monte-carlo tree search: The optimistic principle
  applied to optimization and planning,'' \emph{Foundations and Trends® in
  Machine Learning}, vol.~7, no.~1, pp. 1--129, 2014.

\bibitem{shalev2012online}
S.~Shalev-Shwartz \emph{et~al.}, ``Online learning and online convex
  optimization,'' \emph{Foundations and Trends{\textregistered} in Machine
  Learning}, vol.~4, no.~2, pp. 107--194, 2012.

\bibitem{spall2005introduction}
J.~C. Spall, \emph{Introduction to stochastic search and optimization:
  estimation, simulation, and control}.\hskip 1em plus 0.5em minus 0.4em\relax
  John Wiley \& Sons, 2005, vol.~65.

\bibitem{gokcesu2018density}
K.~Gokcesu and S.~S. Kozat, ``Online density estimation of nonstationary
  sources using exponential family of distributions,'' \emph{{IEEE} Trans.
  Neural Networks Learn. Syst.}, vol.~29, no.~9, pp. 4473--4478, 2018.

\bibitem{willems}
F.~M.~J. Willems, ``Coding for a binary independent
  piecewise-identically-distributed source.'' \emph{IEEE Transactions on
  Information Theory}, vol.~42, no.~6, pp. 2210--2217, 1996.

\bibitem{gokcesu2018anomaly}
K.~Gokcesu and S.~S. Kozat, ``Online anomaly detection with minimax optimal
  density estimation in nonstationary environments,'' \emph{{IEEE} Trans.
  Signal Process.}, vol.~66, no.~5, pp. 1213--1227, 2018.

\bibitem{coding2}
G.~I. Shamir and N.~Merhav, ``Low-complexity sequential lossless coding for
  piecewise-stationary memoryless sources,'' \emph{IEEE Transactions on
  Information Theory}, vol.~45, no.~5, pp. 1498--1519, Jul 1999.

\bibitem{neyshabouri2018asymptotically}
M.~M. Neyshabouri, K.~Gokcesu, H.~Gokcesu, H.~Ozkan, and S.~S. Kozat,
  ``Asymptotically optimal contextual bandit algorithm using hierarchical
  structures,'' \emph{IEEE transactions on neural networks and learning
  systems}, vol.~30, no.~3, pp. 923--937, 2018.

\bibitem{cesa-bianchi}
S.~Bubeck and N.~Cesa{-}Bianchi, ``Regret analysis of stochastic and
  nonstochastic multi-armed bandit problems,'' \emph{Foundations and Trends in
  Machine Learning}, vol.~5, no.~1, pp. 1--122, 2012.

\bibitem{gokcesu2018bandit}
K.~{Gokcesu} and S.~S. {Kozat}, ``An online minimax optimal algorithm for
  adversarial multiarmed bandit problem,'' \emph{IEEE Transactions on Neural
  Networks and Learning Systems}, vol.~29, no.~11, pp. 5565--5580, 2018.

\bibitem{tnnls3}
H.~R. Berenji and P.~Khedkar, ``Learning and tuning fuzzy logic controllers
  through reinforcements,'' \emph{IEEE Transactions on Neural Networks},
  vol.~3, no.~5, pp. 724--740, Sep 1992.

\bibitem{ozkan}
H.~Ozkan, M.~A. Donmez, S.~Tunc, and S.~S. Kozat, ``A deterministic analysis of
  an online convex mixture of experts algorithm,'' \emph{IEEE Transactions on
  Neural Networks and Learning Systems}, vol.~26, no.~7, pp. 1575--1580, July
  2015.

\bibitem{tnnls1}
R.~Song, F.~L. Lewis, and Q.~Wei, ``Off-policy integral reinforcement learning
  method to solve nonlinear continuous-time multiplayer nonzero-sum games,''
  \emph{IEEE Transactions on Neural Networks and Learning Systems}, vol.~PP,
  no.~99, pp. 1--10, 2016.

\bibitem{gokcesu2016prediction}
N.~D. Vanli, K.~Gokcesu, M.~O. Sayin, H.~Yildiz, and S.~S. Kozat, ``Sequential
  prediction over hierarchical structures,'' \emph{IEEE Transactions on Signal
  Processing}, vol.~64, no.~23, pp. 6284--6298, Dec 2016.

\bibitem{singer}
A.~C. Singer and M.~Feder, ``Universal linear prediction by model order
  weighting,'' \emph{IEEE Transactions on Signal Processing}, vol.~47, no.~10,
  pp. 2685--2699, Oct 1999.

\bibitem{tnnls4}
J.~Moody and M.~Saffell, ``Learning to trade via direct reinforcement,''
  \emph{IEEE Transactions on Neural Networks}, vol.~12, no.~4, pp. 875--889,
  Jul 2001.

\bibitem{gokcesu2019outlier}
K.~Gokcesu, M.~M. Neyshabouri, H.~Gokcesu, and S.~S. Kozat, ``Sequential
  outlier detection based on incremental decision trees,'' \emph{{IEEE} Trans.
  Signal Process.}, vol.~67, no.~4, pp. 993--1005, 2019.

\bibitem{gokcesu2016nested}
I.~Delibalta, K.~Gokcesu, M.~Simsek, L.~Baruh, and S.~S. Kozat, ``Online
  anomaly detection with nested trees,'' \emph{{IEEE} Signal Process. Lett.},
  vol.~23, no.~12, pp. 1867--1871, 2016.

\bibitem{gokcesu2017online}
K.~Gokcesu and S.~S. Kozat, ``Online anomaly detection with minimax optimal
  density estimation in nonstationary environments,'' \emph{IEEE Transactions
  on Signal Processing}, vol.~66, no.~5, pp. 1213--1227, 2017.

\bibitem{bartlett2019simple}
P.~L. Bartlett, V.~Gabillon, and M.~Valko, ``A simple parameter-free and
  adaptive approach to optimization under a minimal local smoothness
  assumption,'' in \emph{Algorithmic Learning Theory}.\hskip 1em plus 0.5em
  minus 0.4em\relax PMLR, 2019, pp. 184--206.

\bibitem{grill2015black}
J.-B. Grill, M.~Valko, and R.~Munos, ``Black-box optimization of noisy
  functions with unknown smoothness,'' \emph{Advances in Neural Information
  Processing Systems}, vol.~28, pp. 667--675, 2015.

\bibitem{piyavskii1972algorithm}
S.~Piyavskii, ``An algorithm for finding the absolute extremum of a function,''
  \emph{USSR Computational Mathematics and Mathematical Physics}, vol.~12,
  no.~4, pp. 57--67, 1972.

\bibitem{shubert1972sequential}
B.~O. Shubert, ``A sequential method seeking the global maximum of a
  function,'' \emph{SIAM Journal on Numerical Analysis}, vol.~9, no.~3, pp.
  379--388, 1972.

\bibitem{jacobsen1978global}
S.~E. Jacobsen and M.~Torabi, ``A global minimization algorithm for a class of
  one-dimensional functions,'' \emph{Journal of Mathematical Analysis and
  Applications}, vol.~62, no.~2, pp. 310--324, 1978.

\bibitem{schoen1982sequential}
F.~Schoen, ``On a sequential search strategy in global optimization problems,''
  \emph{Calcolo}, vol.~19, no.~3, pp. 321--334, 1982.

\bibitem{mayne1984outer}
D.~Q. Mayne and E.~Polak, ``Outer approximation algorithm for nondifferentiable
  optimization problems,'' \emph{Journal of Optimization Theory and
  Applications}, vol.~42, no.~1, pp. 19--30, 1984.

\bibitem{mladineo1986algorithm}
R.~H. Mladineo, ``An algorithm for finding the global maximum of a multimodal,
  multivariate function,'' \emph{Mathematical Programming}, vol.~34, no.~2, pp.
  188--200, 1986.

\bibitem{shen1987interval}
Z.~Shen and Y.~Zhu, ``An interval version of shubert's iterative method for the
  localization of the global maximum,'' \emph{Computing}, vol.~38, no.~3, pp.
  275--280, 1987.

\bibitem{horst1987convergence}
R.~Horst and H.~Tuy, ``On the convergence of global methods in multiextremal
  optimization,'' \emph{Journal of Optimization Theory and Applications},
  vol.~54, no.~2, pp. 253--271, 1987.

\bibitem{breiman1993deterministic}
L.~Breiman and A.~Cutler, ``A deterministic algorithm for global
  optimization,'' \emph{Mathematical Programming}, vol.~58, no.~1, pp.
  179--199, 1993.

\bibitem{baritompa1994accelerations}
W.~Baritompa and A.~Cutler, ``Accelerations for global optimization covering
  methods using second derivatives,'' \emph{Journal of Global Optimization},
  vol.~4, no.~3, pp. 329--341, 1994.

\bibitem{hansen1995lipschitz}
P.~Hansen and B.~Jaumard, ``Lipschitz optimization,'' in \emph{Handbook of
  global optimization}.\hskip 1em plus 0.5em minus 0.4em\relax Springer, 1995,
  pp. 407--493.

\bibitem{sergeyev1998global}
Y.~D. Sergeyev, ``Global one-dimensional optimization using smooth auxiliary
  functions,'' \emph{Mathematical Programming}, vol.~81, no.~1, pp. 127--146,
  1998.

\bibitem{ellaia2012modified}
R.~Ellaia, M.~Z. Es-Sadek, and H.~Kasbioui, ``Modified piyavskii’s global
  one-dimensional optimization of a differentiable function,'' \emph{Applied
  Mathematics}, vol.~3, pp. 1306--1320, 2012.

\bibitem{brent2013algorithms}
R.~P. Brent, \emph{Algorithms for minimization without derivatives}.\hskip 1em
  plus 0.5em minus 0.4em\relax Courier Corporation, 2013.

\bibitem{gokcesu2021regret}
K.~Gokcesu and H.~Gokcesu, ``Regret analysis of global optimization in
  univariate functions with lipschitz derivatives,'' \emph{arXiv preprint
  arXiv:2108.10859}, 2021.

\bibitem{horst2013global}
R.~Horst and H.~Tuy, \emph{Global optimization: Deterministic
  approaches}.\hskip 1em plus 0.5em minus 0.4em\relax Springer Science \&
  Business Media, 2013.

\bibitem{danilin1971estimation}
Y.~M. Danilin, ``Estimation of the efficiency of an absolute-minimum-finding
  algorithm,'' \emph{USSR Computational Mathematics and Mathematical Physics},
  vol.~11, no.~4, pp. 261--267, 1971.

\bibitem{malherbe2017global}
C.~Malherbe and N.~Vayatis, ``Global optimization of lipschitz functions,'' in
  \emph{International Conference on Machine Learning}.\hskip 1em plus 0.5em
  minus 0.4em\relax PMLR, 2017, pp. 2314--2323.

\bibitem{bouttier2020regret}
C.~Bouttier, T.~Cesari, and S.~Gerchinovitz, ``Regret analysis of the
  piyavskii-shubert algorithm for global lipschitz optimization,'' \emph{arXiv
  preprint arXiv:2002.02390}, 2020.

\end{thebibliography}
\end{document}